\pdfoutput=1

\documentclass[11pt]{article}

\usepackage{acl}

\usepackage{times}
\usepackage{latexsym}
\usepackage{comment}

\usepackage{mathtools}
\usepackage{bm}
\usepackage{amssymb}

\usepackage[T1]{fontenc}

\usepackage[utf8]{inputenc}

\usepackage{microtype}

\usepackage{inconsolata}

\title{Lifelong Event Detection via Optimal Transport}

\author{Viet Dao\textsuperscript{\rm 1}\thanks{\ \ Equal contribution.}, Van-Cuong Pham\textsuperscript{\rm 1}\footnotemark[1], Quyen Tran\textsuperscript{\rm 1}\footnotemark[1],Thanh-Thien Le\textsuperscript{\rm 1}, \\ {\bf Linh Ngo Van}\textsuperscript{\rm 2} \and {\bf Thien Huu Nguyen}\textsuperscript{\rm 1, 3} \\
  \textsuperscript{\rm 1}VinAI Research \quad \textsuperscript{\rm 2}Hanoi University of Science and Technology \quad \textsuperscript{\rm 3}University of Oregon \\
  \texttt{\{v.vietdt11, v.cuongpv27, v.quyentt15, v.thienlt3\}@vinai.io}, \\
  \texttt{linhnv@soict.hust.edu.vn}, \\
  \texttt{thien@cs.oregon.edu} \\} 

\begin{document}
\maketitle
\begin{abstract}
Continual Event Detection (CED) poses a formidable challenge due to the catastrophic forgetting phenomenon, where learning new tasks (with new coming event types) hampers performance on previous ones. In this paper, we introduce a novel approach, Lifelong Event Detection via Optimal Transport (\textbf{LEDOT}), that leverages optimal transport principles to align the optimization of our classification module with the intrinsic nature of each class, as defined by their pre-trained language modeling. Our method integrates replay sets, prototype latent representations, and an innovative Optimal Transport component. Extensive experiments on MAVEN and ACE datasets demonstrate LEDOT's superior performance, consistently outperforming state-of-the-art baselines. The results underscore LEDOT as a pioneering solution in continual event detection, offering a more effective and nuanced approach to addressing catastrophic forgetting in evolving environments.
\end{abstract}

\section{Introduction}
\label{sec:introduction}
Event Detection (ED) \cite{nguyen-etal-2016-joint-event,nguyen2023retrieving} presents a pivotal challenge in the domain of Information Extraction, tasked with identifying event triggers and their associated types from natural language text.
However, the conventional ED training paradigm, characterized by its static nature, falls short in capturing the dynamic nature of real-world data.
As highlighted by \citet{yu-etal-2021-lifelong}, the ontology of events in ED research has been exhibiting a constant shift since its introduction, prompting the exploration of Continual Event Detection (CED), where data arrives continuously as a sequence of non-overlapping tasks. Although large language models (LLMs) have recently emerged, showcasing the ability to tackle numerous problems using only prompts without the need for fine-tuning, they fall short in the domains of information extraction (IE) \cite{han2023information, gao2023exploring} and continual learning \cite{shi2024continual}. Continual event detection, in particular, remains a difficult task that is not effectively addressed by LLMs.

CED presents many issues, most notably the \emph{catastrophic forgetting} \cite{MCCLOSKEY1989109, Ratcliff1990-RATCMO} phenomenon, where the training signal from new task hampers performance on past tasks. To provide a solution for this issue, numerous methods have been proposed, which usually fall into one of the three eminent approaches: \emph{Regularization-based} \cite{chaudhry2021using, saha2021gradient, phan2022reducing, van2022auxiliary, hai2024continual}; \emph{Architecture-based} \cite{yoon2017lifelong, SOKAR20211}; and \emph{Memory-based} \cite{9009019, NEURIPS2019_fa7cdfad}. Out of these three, Memory-based methods have demonstrated superiority, leveraging access to the \emph{Replay buffer}, a memory of limited size containing a portion of data from previously learned tasks for the model to rehearse during the training of new tasks.

Despite the promise of Memory-based methods, challenges abound. First, the finite capacity of the Replay buffer results in the eviction of valuable information, leading to incomplete representations of past tasks and hence, inadequate generality. Furthermore, the process of sampling and replaying data might not be optimally curated, potentially hindering the model's ability to generalize across tasks effectively.

This setback arises because the conventional practice of discarding the original head of pre-trained language models (PLMs) during fine-tuning on downstream tasks overlooks valuable linguistic information encoded within it. In training the classifier module, state-of-the-art approaches \cite{qin2024lifelong, wang2023continual, liu-etal-2022-incremental, yu-etal-2021-lifelong} often do so in isolation, devoid of any priors or foundations. Discarding the language modeling head in PLMs is highly wasteful. The language modeling head contains essential information about vocabulary distribution based on contextual representations. Losing this head sacrifices crucial linguistic nuances, making it harder to align the classifier module and ensure efficient fine-tuning. Aligning our classifier module to this information is an essential but also formidable challenge.  This alignment is crucial for ensuring a more efficient fine-tuning process, as it provides a foundational standard of learning that mitigates unnecessary overplasticity and prevents catastrophic forgetting.

To address the limitations discussed, this paper introduces a method to enhance Memory-based CED by integrating Optimal Transport (OT) principles, which provide a robust framework for measuring the distance between probability distributions. By incorporating OT into the fine-tuning process, we aim to retain essential linguistic information from the PLM head, ensuring the model remains invariant to specific tasks. This integration involves defining an appropriate cost matrix, a key challenge that we address by proposing a novel construction tailored to our method. Our approach ensures effective alignment between the PLM head and the classifier's output, leveraging OT to enhance the model's performance and robustness across various tasks while preserving the PLM's inherent linguistic knowledge.

\section{Background}
\label{sec:background}

\subsection{Event Detection}
\label{sec:background:ed}
Following previous works, we formalize Event Detection as a Span-based Classification task \cite{nguyen-grishman-2015-event,lu-nguyen-2018-similar,man-duc-trong-etal-2020-introducing}. Given an input instance $x=(w_{1:L}, s, e)$ consisting of a $L$-token context sequence $w_{1:L}$, a start index $s$, and an end index $e$, an ED model has to learn to assign the text span $w_{s:e}$ into a label $y$ from a set of pre-defined event types $\mathcal{Y}$, or \texttt{NA} if $w_{s:e}$ does not trigger a known event.

Generally, we use a language model $\mathcal{M}$ to encode the context sequence $w_{1:L}$ into contextualized representation $w'_{1:L}$. Then, a classifier is utilized to classify the representation of the trigger span:
\begin{equation}
    \centering
    h = [w'_s, w'_e],
    \label{eq:2}
\end{equation}
\begin{equation}
    \centering
    p(y|x) = Softmax(Linear(FNN(h)).
    \label{eq:3}
\end{equation}
Here, $FNN$ denotes a feed-forward neural network, $[\cdot,\cdot]$ is the concatenation operation, $h$ is the hidden vector representing $w_{s:e}$, and $p(y|x)$ models the probability of predicting $y$ from the input $x$.

The model is trained on a dataset $\mathcal{D}=\{(x_i, y_i)\}_{i=1}^N$ using cross-entropy loss:
\begin{equation}
    \centering
    \mathcal{L}_C(\mathcal{D}) = -\frac{1}{\vert \mathcal{D} \vert} \sum_{(x, y) \in \mathcal{D}} \log p(y|x).
    \label{eq:4}
\end{equation}

To mitigate the imbalance between the number of event triggers and the number of \texttt{NA} spans, we re-weight the loss with a hyperparameter $\eta$:
\begin{equation}
    \centering
    \mathcal{L}_C = \eta\mathcal{L}_C(\mathcal{D_\texttt{NA}}) + (1-\eta)\mathcal{L}_C(\mathcal{D} \setminus \mathcal{D_\texttt{NA}})
    \label{eq:5}
\end{equation}
where $\mathcal{D_\texttt{NA}}$ is the set of \texttt{NA} instances.

\subsection{Continual Event Detection}
\label{sec:background:ced}
The training data in CED is not static but arrives sequentially as a stream of $T$ non-overlapping tasks $\{\mathcal{D}_t | \bigcup_{t=1}^{T}\mathcal{D}_t=\mathcal{D}; \mathcal{D}_t \cap \mathcal{D}_{t'} = \emptyset \}$. At each timestep $t$, the $t^{th}$ task data only covers a set of event types $\mathcal{Y}_t = \{y_t^1, y_t^2, \ldots y_t^{n_t}\}$, which is a subset of the full ontology of event types $\mathcal{Y}$. Here, unseen events and negative instances (i.e. text spans that do not trigger any event) are treated as \texttt{NA}. After training on $\mathcal{D}_t$, the model is expected to be able to detect all seen events thus far, i.e. $\mathcal{Y}_1 \bigcap \mathcal{Y}_2 \ldots \bigcap \mathcal{Y}_t$. To this end, we employ two commonly used techniques in Rehearsal-based Continual Learning: \emph{Naive Replay}, and \emph{Knowledge Distillation} \citep{hinton2015distilling}. Let $R_{t-1}$ be the replay buffer up to task $t-1$, the Replay Loss and Knowledge Distillation loss are written as follows:
\begin{equation}
    \centering
    \mathcal{L}_R = - \frac{1}{|\mathcal{R}_{t-1}|}\sum_{(h, y) \in R_{t-1}} \log p^t(y|h),
    \label{eq:5.5}
\end{equation}
\begin{equation}
    \centering
    \mathcal{L}_D = - \sum_{(h, y) \in R_{t-1}} p^{t-1}(y|h) \log p^t(y|h),
    \label{eq:6}
\end{equation}
where $p^t$ denotes the probability of predictions given by the model instance at timestep $t$.

\label{sec:methodology}

\section{Lifelong Event Detection via Optimal Transport}
\label{sec:methodology:ot}
We incorporate Optimal Transport (OT) as a foundational element of our methodology. OT is a mathematical framework designed to compute the distance between two probability distributions with different supports.

In our methodology, OT is applied to align the probability distribution output of the classifier head with the distributional characteristics inherent in the vocabulary of the Pre-trained Language Model (PLM) head. The softmax class probabilities from the classifier head are transported to closely match the pre-trained distribution, facilitating a seamless integration of task-specific knowledge while minimizing the divergence from the model's pre-existing linguistic understanding.

We forward each event trigger through a pre-trained language model and its original language modeling head, and obtain a distribution over a dictionary of $V$ words: 
\begin{align*}
    x_s &= Softmax(LMH(w'_s)/\tau) \\
    x_e &= Softmax(LMH(w'_e)/\tau) \\
    \tilde x &= (x_s+ x_e)/2
\end{align*}
where LMH is a pre-trained language model head, \(\tau\) is temperature coefficient, and \(\tilde x\) is distribution of the event trigger over dictionary. 

Each event trigger is associated with a distribution over $C$ classes: $\bm{p} \in \Delta^C$, where each entry indicates the probability that the event trigger belongs to a class in the ontology. An encoder is employed to generate $\bm{p}$ from $\bm{x}$, defined as $\bm{p} = Softmax(\theta(x))$, where $\theta$ represents the parameters of the neural network as described in Section \ref{sec:background:ed}.

Given that $\bm{\tilde x}$ and $\bm{p}$ are distributions with different supports for the same event trigger, we aim to train the model by minimizing the following Optimal Transport (OT) distance to push $\bm{p}$ towards $\bm{\tilde x}$:
\begin{equation} \label{eq:8}
d_{\bm{\mathrm{M}}}(\bm{\tilde x}, \bm{p}) \coloneqq \underset{\mathbf{P} \in U(\bm{\tilde x}, \bm{p})}{\rm min} \ \langle\mathbf{P}, \mathbf{M}\rangle,
\end{equation}
where $\langle\cdot, \cdot\rangle$ denotes the Frobenius inner product; the cost matrix $\bm{\mathrm{M}} \in \mathbb{R}{\geq0}^{V \times C}$ captures semantic distances between class $c$ and word $v$, with each entry $m_{vc}$ signifying the importance of words in the corresponding class; $\mathbf{P} \in \mathbb{R}_{>0}^{V \times C}$ denotes the transport plan; and and $U(\bm{\tilde x}, \bm{p})$ is defined as the set of all viable transport plans. Considering two discrete random variables $X \sim \mathrm{Categorical}(\bm{\tilde x})$ and $Y \sim \mathrm{Categorical}(\bm{p})$, where the transport plan $\mathbf{P}$ becomes a joint probability distribution of $(X, Y)$, i.e., $\mathrm{p}(X=i, Y=j) = p_{ij}$: the set $U(\bm{\tilde x}, \bm{p})$ encompasses all possible joint probabilities that satisfy the specified constraints, forming a transport polytope.

Directly optimizing Eq. (\ref{eq:8}) poses a time-consuming challenge. To address this, an entropic-constrained regularized optimal transport (OT) distance is introduced, known as the Sinkhorn distance:
\begin{equation} \label{eq:8.1}
s_{\bm{\mathrm{M}}}(\bm{\tilde x}, \bm{p}) \coloneqq \underset{\mathbf{P} \in U(\bm{\tilde x}, \bm{p})}{\rm min} \ \langle\mathbf{P}, \mathbf{M}\rangle - \mathbf{H(P)},
\end{equation}
where the entropy function of the transport plan $\mathbf{H(P)} \stackrel{\text{def}}{=} - \sum_{i, j}\mathbf{P}_{i, j}(\log(\mathbf{P}_{i, j} - 1))$ is the regularizing function
\citep{NIPS2013_af21d0c9}.

The cost matrix $\bm{\mathrm{M}}$ is a trainable variable in our model. To overcome the challenge of learning the cost function, we propose a specific construction for $\bm{\mathrm{M}}$:
\begin{equation} \label{eq:9}
m_{vc} = 1 - \mathrm{cos}(\mathbf{e}_v, \mathbf{g}_c),
\end{equation}
where $\mathrm{cos}(\bm{\cdot}, \bm{\cdot})$ represents the cosine similarity, and $\mathbf{g}_c \in \mathbb{R}^D$ and $\mathbf{e}_v \in \mathbb{R}^D$ are the embeddings of class $c$ and word $v$, respectively. After training on one task, the learned class embeddings are frozen. We then expand the size of the class embeddings and train the newly initialized embeddings on the new task.

\citet{frogner2015learning} further suggested combining the OT loss with a conventional cross-entropy loss to better guide the model. By parameterizing $\bm{M}$ with $\bm{G}$, the collection of class embeddings, the final OT objective function is expressed as:
\begin{equation} \label{eq:10}
\mathcal{L_{OT}} = \underset{\theta, \bm{\mathrm{G}}}{\rm min} \ [s_{\bm{\mathrm{M}}}(\bm{\tilde x}, \bm{p}) - \epsilon\bm{\tilde x} \ \mathrm{log} \phi(\bm{p})].
\end{equation}

To maintain the consistency of class representations across tasks, an additional regularization term enforces the proximity of class representations in the current task to those in the most recent task:
\begin{equation}
\mathcal{L}_G = ||\mathbf{G}_{t} - \mathbf{G}_{(t-1)}||^2.
\end{equation}
Finally, we can write our final objective function:
\begin{equation}
\mathcal{L} = \mathcal{L}_C + \mathcal{L}_R + \mathcal{L}_D + \mathcal{L}_{OT} + \alpha\mathcal{L}_G,
\end{equation}
where $\alpha$ is the regularization coefficient.


\begin{table*}[t]
\centering
\resizebox{\linewidth}{!}{
    \begin{tabular}{l | c c c c c | c c c c c} 
    \hline
     & \multicolumn{5}{c|}{\textbf{MAVEN}} & \multicolumn{5}{c}{\textbf{ACE}} \\
  
    Task & 1 & 2 & 3 & 4 & 5 & 1 & 2 & 3 & 4 & 5 \\

    \hline
    BIC &63.16&55.51&53.96&50.13&49.07& 55.88&58.16&61.23&59.72&59.02 \\
    KCN & 63.16 & 55.73 & 53.69 & 48.86 & 47.44 & 55.88 & 58.55 & 61.40 & 59.48 & 58.64 \\
    KT & 62.76 & 58.49 & 57.46 & 55.38 & 54.87 & 55.88 & 57.29 & 61.42 & 60.78 & 59.82 \\
    EMP &  66.82 & 58.02 & 58.19 & 55.07 & 54.52 & 59.05 & 57.14 & 55.80 & 53.42 & 52.97\\
    ESCO &  67.50 &	 61.37& 	60.65& 	57.43& 	57.35& ---- & ---- & ---- & ---- & ---- \\
    SCR & \bf 76.52	&57.97	&57.89	&52.74&	53.41&\bf75.24	&\bf63.3	&61.07	&55.05	&55.37\\
    SharpSeq&62.28&	61.85&	62.92&	61.31&	60.27&56.47&	56.99&	64.44&	62.47&	62.60\\

    \hline
    
    LEDOT-OT & 63.34 & 59.89 & 59.28 & 56.24 & 55.20 & 58.74 & 58.08 & 61.81 & 58.32 & 59.76 \\
    LEDOT-R & 63.01 & 60.16 & 59.76 & 56.75 & 54.59	& 58.30 & 58.60 & 63.14 & 58.82 & 60.18 \\
    LEDOT-P & 63.01 & 59.95 & 59.32 & 56.10 & 55.21 &  59.95 &  56.63 &  62.09 & 60.08 & 61.41 \\
    LEDOT & 62.98 &  60.47 &  60.78 & 58.53 & 57.53 & 58.30 & 59.69 & 63.52 & 61.05 & 63.22 \\
    LEDOT + SharpSeq&63.30 &\bf 61.97&	\bf63.00&	\bf61.81&	\bf61.49&60.15&59.73 &	\bf64.55&	\bf63.65&	\bf64.27\\
    \hline
    \it Upperbound & \slash & \slash & \slash & \slash & \it 64.14 & \slash & \slash & \slash & \slash & \it 67.95 \\
    
    \hline
    \end{tabular}
    }
    \caption{\label{tab:show-rs}
    Classification F1-scores (\%) on 2 datasets MAVEN and ACE. \textit{Upperbound} indicates the theoretical maximum achievable performance when BERT is frozen. 
    }
\end{table*}

\paragraph{Avoiding Catastrophic Forgetting}
Similar to many CED baselines, our method incorporates a replay process. However, our approach to constructing the memory buffer is distinct. For each class in the training data, we retain the prototype mean $\mu$ and diagonal covariance $\Sigma$ of its trigger representations encountered by the model, rather than storing explicit data samples. During replay, synthetic samples are generated from these prototypes and combined with the replay buffer $\mathcal{R}$ to form the effective buffer $\tilde{\mathcal{R}}$. This modified buffer replaces $\mathcal{R}$ in the computation of $\mathcal{L}_R$ (\ref{eq:5.5}) and $\mathcal{L}_D$ (\ref{eq:6}).


\section{Experiments}
\label{sec:exp}

\subsection{Settings}
\label{sec:exp:settings}

\paragraph{Datasets} We employ two datasets in our experiments: ACE 2005 \cite{walker2006ace} and MAVEN \cite{wang2020maven}; both are preprocessed similar to \citeposs{yu-etal-2021-lifelong} work.
To ensure fairness, we rerun all baselines on the same preprocessed datasets.
The detailed statistics of the two datasets can be found in Appendix \ref{sec:appendix:datasets}.

\paragraph{Experimental Settings} We adopt the Oracle negative setting, as mentioned by \citet{yu-etal-2021-lifelong}, to evaluate all methods in continual learning scenario. This setting involves excluding the learned types from previous tasks in the training set of the new task, except for the \texttt{NA} (Not Applicable) type. Labels for future tasks are treated as \texttt{NA} type. Assessments are conducted using the exact same task permutations as in \citeposs{yu-etal-2021-lifelong} work. The performance metric is the average terminal F1 score across 5 permutations after each task. Recently, \cite{le2024sharpseq} introduced a multi-objective optimization method that is compatible with our proposed LEDOT approach. To examine the impact of LEDOT on SharpSeq, we conducted an experiment referred to as LEDOT+SharpSeq. For details on other baselines and the integration of LEDOT with SharpSeq, please refer to Appendix \ref{sec:appendix:baseline}.

\subsection{Experimental Results}
\label{sec:exp:results}

Table \ref{tab:show-rs} showcases the impressive results of our proposed method, \textbf{LEDOT}, compared to state-of-the-art baselines in continual event detection. On both the MAVEN and ACE datasets, LEDOT consistently achieves higher F1 scores, surpassing most baseline methods. When combined with SharpSeq, LEDOT further enhances performance, increasing the F1-score by a significant margin of $1.22\%$ on MAVEN and $1.67\%$ on ACE after five tasks. 

We also conduct further ablation studies to evaluate variants of \textbf{LEDOT}: LEDOT-OT (without Optimal Transport), LEDOT-R (without the replay set), and LEDOT-P (without prototype latent representations). Even without prototype rehearsal, LEDOT-P with OT surpasses the replay-based baseline KT by $0.34\%$ on MAVEN and $1.59\%$ on ACE. Moreover, LEDOT outperforms LEDOT-OT, highlighting the crucial role of OT in preventing catastrophic forgetting. Specifically, OT improves F1 scores by $2.33\%$ on MAVEN and $3.46\%$ on ACE. These results emphasize the importance of OT in mitigating catastrophic forgetting in continual event detection.


\section{Conclusion}
Harnessing the inherent linguistic knowledge from pre-trained language modeling heads in encoder-based language models play a pivotal role in enhancing performance in downstream tasks. With the introduction of LEDOT, we present a novel approach utilizing optimal transport to align the learning of each task with a common reference—the pre-trained distribution of the vocabulary. This alignment mitigates overfitting to the current task and effectively addresses the challenge of catastrophic forgetting. Our method, demonstrating superior performance across various benchmarks, stands as a testament to the effectiveness of leveraging pre-trained language modeling heads for continual event detection, offering a promising avenue for future research in this domain. In the future, we plan to extend our method to solve continual learning challenges for other information extraction tasks, such as event-event relation extraction \cite{man2024hierarchical,man-etal-2024-hierarchical}.

\section*{Limitations}
Being an empirical study into the effectiveness of Optimal Transport in aligning the output distribution of Continual Event Detection models, our work is not without limitations. We acknowledge this, and would like to discuss our limitations as follows:

\begin{itemize}
    \item The method proposed in this paper is orthogonal to the tasks of interest and the specific techniques to solve them. With that being said, our method is applicable to a wide range of information extraction tasks, such as Named Entity Recognition, and Relation Extraction, as well as other text classification tasks, such as Sentiment Analysis. However, given limited time and computational resources, we limit the scope of our experiments to only Event Detection. The extent to which our proposed method can work with other NLP problems can be an interesting topic that we leave for future work. Nevertheless, our experimental results suggest that using Optimal Transport to align the output distribution of the model with the pre-trained language modeling head has the potential to improve continual learning performance on other problems as well.
    \item This paper presents the empirical results of our LEDOT method using a pre-trained encoder language model (i.e. BERT) as the backbone. Meanwhile, large decoder-only language models, with their heavily over-parameterized architectures, amazing emergent ability, and great generalization capability, have emerged and become the center of focus of NLP research in recent years. Though they have proved to be able to understand language and solve almost all known NLP tasks without needing much fine-tuning, many studies (\citep{lai-etal-2023-chatgpt}; \citep{QIU2024200308}; \citep{zhong2023chat}) suggested that even the largest models like ChatGPT (\citep{NEURIPS2022_b1efde53}) still lag behind smaller but specialized models such as BERT \citep{devlin-etal-2019-bert} and T5 \citep{raffel2023exploring} by a significant margin on tasks like Event Detection. We thus believe that studies on the applications of encoder language models in Continual Event Detection are still needed.
\end{itemize}


\section*{Acknowledgements}

This research has been supported by the Army Research Office (ARO) grant W911NF-21-1-0112, the NSF grant CNS-1747798 to the IUCRC Center for Big Learning, and the NSF grant \# 2239570. This research is also supported in part by the Office of the Director of National Intelligence (ODNI), Intelligence Advanced Research Projects Activity (IARPA), via the HIATUS Program contract 2022-22072200003. The views and conclusions contained herein are those of the authors and should not be interpreted as necessarily representing the official policies, either expressed or implied, of ODNI, IARPA, or the U.S. Government.

\bibliography{anthology,custom}

\clearpage

\appendix

\section{Additional Experimental Details}

\subsection{Baselines}
\label{sec:appendix:baseline}

The following continual learning and continual ED methods are employed as baselines in this paper:
\begin{itemize}
    
    \item \textbf{BIC} \cite{wu2019large} addresses model bias towards new labels via an affine transformation.
    
    \item \textbf{KCN} \cite{cao2020incremental} employs a limited set to store data for replay, utilizing knowledge distillation and prototype-enhanced retrospection to alleviate catastrophic forgetting. \item \textbf{KT} \cite{yu-etal-2021-lifelong} follows a memory-based approach, combining knowledge distillation with knowledge transfer. This method utilizes new-label samples to reinforce the model's retention of old knowledge and employs old-label samples to initialize representations for new-label data in the classification layer. 
    \item \textbf{EMP} \cite{liu-etal-2022-incremental} also leverages knowledge distillation and introduces straight prompts into the input text to retain previous knowledge.
    \item \textbf{ESCO} \cite{qin2024lifelong} introduce ESCO, a method combining Embedding Space Separation and Compaction. ESCO pushes the feature distribution of new data away from old data to reduce interference and pulls memory data towards its prototype to improve intra-class compactness and alleviate overfitting on the replay dataset. 
    \item \textbf{SharpSeq} The framework introduced in \textbf{SharpSeq} \cite{le2024sharpseq} integrates multi-objective optimization (MOO) with sharpness-aware minimization (SAM). In the context of continual learning, handling multiple losses often involves simply summing them with fixed coefficients. However, this approach can lead to gradient conflicts that hinder the discovery of optimal solutions. MOO algorithms address this issue by dynamically estimating coefficients based on the gradients of the losses. To refine the results of MOO, \cite{le2024sharpseq} employs SAM to identify flat minima along the Pareto front.
    \item \textbf{SCR} \cite{wang2023continual} employs a training approach involving both BERT and the classifier layer. Initially, this yields high F1 scores on early tasks, but performance deteriorates rapidly as more tasks are encountered. In contrast, our method maintains BERT's parameters fixed during training. The SCR approach, which fine-tunes BERT, presents challenges for continual event detection. Despite having different label sets, many sentences are recurrent across tasks. SCR tackles this by using pseudo labels from the previous stage to predict labels on new datasets, containing sentences from previous tasks. However, this strategy leads to data leakage from old tasks to new ones, significantly inflating SCR's replay dataset beyond what is allowed in strict continual learning setups. In contrast, our method relies on a frozen BERT for feature extraction, ensuring consistency in trigger representations over time. Our approach aligns with the principles of continual learning, where the model solely accesses data relevant to the current task. 
    Moreover, the evaluation metric in SCR differs from our approach, as they do not account for the NA label despite it being the most common label in these datasets. Therefore, we have reproduced the results and reported them in Table\ref{tab:show-rs}.

    \item \textbf{LEDOT + SharpSeq} Our proposed method incorporates two key objectives: one focusing on the OT loss for the language modeling head and another serving as a regularization term to ensure the proximity of class representations. Instead of treating these objectives as separate entities within a multi-objective optimization algorithm, we integrate them directly into the overall loss calculation using the same data. This approach maintains the original number of losses, streamlining the optimization process.
\end{itemize}

\subsection{Datasets}
\label{sec:appendix:datasets}
Detailed statistics regarding the datasets used for all empirical assessments can be found in Table \ref{tab:data-spec}.

\subsection{Implementation Details}
\label{sec:appendix:impdetails}
In our experiments, the encoder and language model head is taken from BERT-large-cased \cite{devlin-etal-2019-bert} and they are freezed in the training process. We employ the AdamW optimizer \cite{loshchilov2017decoupled} with a learning rate of $1 \times 10^{-4}$ and a weight decay of $1 \times 10^{-2}$. Model training continues until there is no increase in performance on the development set. The replay setting remains consistent with KT \citeposs{yu-etal-2021-lifelong}, where the number of instances for each label in the replay set is set to 20. Since the size of the vocabulary is large and it contains many subwords and completely unrelated words, to reduce the computation, we select only a subset of words that are verbs. In each batch, we combine that set with tokens in the batch to compute the OT loss.

All implementations are coded using PyTorch, and the experiments were carried out on NVIDIA A100 and NVIDIA V100 GPUs.

\section{Ablation Study}
\label{sec:appendix:ablation}

\subsection{Temperature of Language Modeling Head}
\label{sec:appendix:lmheadtemp}

We conduct an ablation study to explore the impact of different temperatures in the language modeling head within the LEDOT method. The motivation behind this study lies in the stochastic nature of the language modeling process, where a higher temperature introduces more randomness. This increased stochasticity can influence the generation not only of the primary label (event type) but also of other words related to the topic. By systematically varying the temperature parameter, denoted as $\tau$, we aim to understand how these different levels of stochasticity affect LEDOT's performance. The results are presented in Table \ref{tab:show-rs-temp}.

\subsection{Quantity of Generated Samples}
\label{sec:appendix:genratio}

In Table \ref{tab:show-rs}, we observe that the performance of LEDOT significantly improves when synthesizing representations from prototypes. To further investigate this effect, we conducted additional experiments with LEDOT, varying the ratios ($r$) between the number of generated samples and the replay set. The outcomes for four $r$ values are presented in Table \ref{tab:show-rs-gen}. Notably, on MAVEN, the highest performance is achieved with $r=10$, yielding a $57.53\%$ F1 score in the fifth task. Conversely, for the fifth task on ACE, the optimal $r$ value is 2020, resulting in a $63.22\%$ score. The influence of prototype sampling on early tasks is relatively marginal, but it becomes more pronounced in later tasks. It is important to note that an increased $r$ value does not necessarily guarantee improved LEDOT performance. This can be attributed to the noise introduced by random processes during representation sampling. The noise can impact the outcome of the language modeling head in LEDOT and potentially misguide the classification head during model optimization. Therefore, when generating more samples, careful consideration is required to mitigate noise effects and avoid adversarial impacts.

\subsection{Further Analysis}
We conduct additional ablation studies to gain deeper insights into the performance of LEDOT. First, we compare the impact of two different initialization methods for Optimal Transport—random initialization and initializing labels by mapping them to their corresponding word embeddings in the vocabulary. The results of this comparison are detailed in Table \ref{tab:show-rs-oti_nit}, shedding light on the influence of initialization strategies on the overall effectiveness of LEDOT. Second, we explore the sensitivity of our method to the coefficient of regularization applied to the cross-task class representations. The results of this investigation are presented in Table \ref{tab:show-rs-rgz}, providing valuable information about the robustness of LEDOT to variations in the regularization coefficient. These ablation studies contribute to a comprehensive understanding of the factors influencing LEDOT's performance in continual event detection scenarios.

\section{Optimal Transport on Continual Relation Extraction}

Our proposed Optimal Transport alignment extends beyond Continual Event Detection: it can also enhance other continual NLP solutions utilizing various kinds of pre-trained language models. To substantiate this claim, we demonstrate its effectiveness in Continual Relation Extraction (CRE) \cite{han-etal-2020-continual, cui-etal-2021-refining, zhao-etal-2022-consistent, xiong2023rationale,nguyen2023spectral,le2024continual} using an \emph{encoder-decoder} language model, specifically T5 \cite{2020t5}.


Our experiments are centered around the state-of-the-art CRE baseline RationaleCL \cite{xiong2023rationale}. This method leverages rationales generated by \texttt{ChatGPT-3.5}\footnote{https://chat.openai.com/} during training to enhance the T5 model for CRE. RationaleCL operates by first generating rationales for current relation samples using an LLM. These rationales are then integrated into the original training dataset for multi-task rationale tuning. Formally, RationaleCL introduces three key objectives:
\begin{align}
    Task_c: & \ x_i \longrightarrow y_i \\
    Task_r: & \ x_i \longrightarrow r_i + y_i \\
    Task_d: & \ x_i + r_i \longrightarrow y_i
\end{align}
where $x_i$ represents the input text, $y_i$ denotes the relation label, and $r_i$ stands for the rationale.
$Task_c$ directly generates the label $y_i$ from the input $x_i$, while $Task_r$ requires the model to generate an explanation before generating an answer. $Task_d$ uses both the input and rationale in the encoder part to answer the question. It is noteworthy that, similar to most continual learning methods, RationaleCL employs a replay process. This process trains the model on both newly encountered data and a limited amount of samples from previously encountered tasks stored in the buffer.



The state-of-the-art performance achieved by RationaleCL in CRE underscores its efficacy. However, our integration of Optimal Transport (OT) methodologies aims to elevate the method to new heights. We introduce OT objectives to align the learned language-modeling head with T5's original language-modeling head, resulting in the enhancements observed over the baseline on the TACRED dataset \cite{zhang2017tacred}, as showcased in Table \ref{tab:oncre}.

Our integration of OT objectives not only mitigates the detrimental effects of catastrophic forgetting but also emerges as a compelling solution for enhancing the fine-tuning process across various downstream tasks.

\begin{table*}
\centering
    \begin{tabular}{ c | c c c c  | c c c c} 
    \hline
     & \multicolumn{4}{c}{\textbf{MAVEN}} & \multicolumn{4}{c}{\textbf{ACE}} \\
  
     & \#Doc & \#Sentence & \#Mention & \#Negative & \#Doc & \#Sentence & \#Mention & \#Negative \\

    \hline
    Train & 2522 & 27983 & 67637 & 280151 & 501 & 18246  & 4088 & 261027\\
    Dev  & 414 & 4432 & 10880 & 46318 & 41 & 1846 &  433 & 53620 \\
    Test & 710 & 8038 & 18904 & 79699  & 55 & 689 & 790 & 93159 \\
    \hline
    \end{tabular}
    \caption{\label{tab:data-spec}
    Statistics of two datasets. \#Doc stands for the total number of documents.
    }
\end{table*}

\begin{table*}
    \centering
    \captionsetup{justification=centering}
    \begin{tabular}{l | c c c c c | c c c c c} 
    \hline
    & \multicolumn{5}{c}{\textbf{MAVEN}} & \multicolumn{5}{c}{\textbf{ACE}} \\
    Task & 1 & 2 & 3 & 4 & 5 & 1 & 2 & 3 & 4 & 5 \\
    \hline
    $\tau = 5$ & \bf 63.15 & \bf 60.78 & 60.66 & 58.51 & 57.37 & 57.41 & 59.00 & 63.60 & 60.87 & 61.81 \\
    $\tau = 4$ & 63.08 & 60.72 & 60.71 & \bf 58.76 & \bf 57.71 & 61.09 & \bf 60.12 & 63.36 & 61.09 & 61.15 \\
    $\tau = 3$ & 63.06 & 60.77 & 60.70 & 58.43 & 57.30 & 58.09 & 59.46 & 63.98 & \bf 61.63 & 62.36 \\
    $\tau = 2$ & 63.11 & 60.64 & 60.70 & 58.45 & 57.50 & 58.30 & 59.69 & 63.52 & 61.05 & \bf 63.22 \\
    $\tau = 1$ & 62.98 & 60.47 & \bf 60.78 & 58.53 & 57.53 & 60.42 & 59.76 & \bf 64.28 & 61.52 & 62.84 \\
    $\tau = 0.1$ & 62.52 & 60.31 & 60.51 & 58.31 & 57.13 & 61.51 & 57.01 & 62.94 & 60.18 & 61.22 \\
    $\tau = 0.01$ & 62.60 & 60.3 & 60.43 & 58.22 & 57.15 & \bf 62.15 & 57.08 & 63.51 & 59.48 & 61.29 \\
    \hline
    \end{tabular}
    \caption{\label{tab:show-rs-temp}
    Ablation results for the temperature of the language modeling head in the LEDOT method.
    }
\end{table*}

\begin{table*}
    \centering
    \captionsetup{justification=centering}
    \begin{tabular}{l | c c c c c | c c c c c} 
    \hline
    & \multicolumn{5}{c}{\textbf{MAVEN}} & \multicolumn{5}{c}{\textbf{ACE}} \\
    Task & 1 & 2 & 3 & 4 & 5 & 1 & 2 & 3 & 4 & 5 \\
    \hline
    $r = 20$ & 63.01 & 60.12 & 60.26 & 57.96 & 56.87 & 58.30 & 59.69 & 63.52 & 61.05 & \bf 63.22 \\
    $r = 10$ & 62.98 & \bf 60.47 & \bf 60.78 & \bf 58.53 & \bf 57.53 & 58.30 & \bf 60.80 & \bf 64.63 & \bf 62.47 & 62.63 \\
    $r = 5$ & 63.01 & 60.30 & 60.54 & 58.22 & 57.01 & 58.30 & 61.06 & 64.67 & 60.59 & 62.29 \\
    $r = 1$ & \bf 63.07 & 60.16 & 60.00 & 57.07 & 55.84 & 58.30 & 60.51 & 64.24 & 60.15 & 62.18 \\
    \hline
    \end{tabular}
    \caption{\label{tab:show-rs-gen}
    Ablation results for the number of generated representations in the LEDOT method.
    }
\end{table*}

\begin{table*}
    \centering
    \captionsetup{justification=centering}
    \begin{tabular}{l | c c c c c | c c c c c} 
    \hline
    & \multicolumn{5}{c}{\textbf{MAVEN}} & \multicolumn{5}{c}{\textbf{ACE}} \\
    Task & 1 & 2 & 3 & 4 & 5 & 1 & 2 & 3 & 4 & 5 \\
    \hline
    random & \bf 63.15 & \bf 60.78 & 60.66 & 58.51 & 57.37 & 57.41 & 59.00 & \bf 63.60 & 60.87 & \bf 61.81 \\
    mapping & 63.08 & 60.72 & \bf 60.71 & \bf 58.76 & \bf 57.71 & \bf 61.09 & \bf 60.12 & 63.36 & \bf 61.09 & 61.15 \\
    \hline
    \end{tabular}
    \caption{\label{tab:show-rs-oti_nit}
    Ablation results for the initialization of Optimal Transport in the LEDOT method. "mapping" indicates initializing labels by mapping them to their corresponding word embeddings in the vocabulary.
    }
\end{table*}

\begin{table*}
    \centering
    \captionsetup{justification=centering}
    \begin{tabular}{l | c c c c c | c c c c c} 
    \hline
    & \multicolumn{5}{c}{\textbf{MAVEN}} & \multicolumn{5}{c}{\textbf{ACE}} \\
    Task & 1 & 2 & 3 & 4 & 5 & 1 & 2 & 3 & 4 & 5 \\
    \hline
    $\alpha = 1$ & 63.01 & 60.36 & 60.67 & 58.33 & 57.41 & 58.30 & 59.72 & 64.41 & 60.97 & 62.29 \\
    $\alpha = 0.5$ & 62.98 & 60.47 & \bf 60.78 & \bf 58.53 & \bf 57.53 & 58.30 & 59.69 & 63.52 & 61.05 & \bf 63.22 \\
    $\alpha = 0.2$ & \bf 63.07 & \bf 60.66 & 60.67 & 58.37 & 57.16 & 58.72 & 59.39 & 64.55 & 61.88 & 62.68 \\
    $\alpha = 0.1$ & 63.01 & 60.45 & 60.60 & 57.79 & 57.02 & 58.72 & \bf 60.01 & \bf 64.61 & \bf 62.49 & 62.82 \\
    \hline
    \end{tabular}
    \caption{\label{tab:show-rs-rgz}
    Ablation results for regularization on cross-task class representations in the LEDOT method.
    }
\end{table*}

\begin{table*}
\centering
    \begin{tabular}{l | c c c c c c c c c c} 
    \hline
     & \multicolumn{10}{c}{\textbf{TACRED}} \\
  
    Task & 1 & 2 & 3 & 4 & 5 & 6 & 7 & 8 & 9 & 10 \\

    \hline
    RCL& 100.00 & 94.80 & 92.20 & 89.24 & 86.56 & 84.74 & 80.57 & 77.46 & 80.98 & 79.11 \\
    OT RCL& 97.76 & 98.40 & 93.17 & 87.94 & 90.18 & 86.05 & 82.73 & 80.61 & 82.61 & 79.36 \\
    \hline
    \end{tabular}
    \caption{\label{tab:oncre}
    Classification accuracy (\%) on the TACRED dataset. RCL abbreviates for RationaleCL.
    }
\end{table*}

\end{document}